\documentclass[conference]{IEEEtran}
\IEEEoverridecommandlockouts
\usepackage{cite}
\usepackage[caption=false,font=tiny,labelfont=rm,textfont=rm]{subfig}
\usepackage{longtable}
\usepackage{amsmath}
\usepackage{amssymb}
\usepackage{multirow}
\usepackage{verbatim}
\usepackage{graphicx}
\usepackage{utfsym}

\def\BibTeX{{\rm B\kern-.05em{\sc i\kern-.025em b}\kern-.08em
    T\kern-.1667em\lower.7ex\hbox{E}\kern-.125emX}}
\begin{document}

\title{Addressing Noise and Stochasticity in Fraud Detection for Service Networks}
%
%
\author{

\IEEEauthorblockN{
Wenxin Zhang \IEEEauthorrefmark{1}, 
Ding Xu\IEEEauthorrefmark{2},
Xi Xuan \IEEEauthorrefmark{3},
Lei Jiang \IEEEauthorrefmark{1},
Guangzhen Yao \IEEEauthorrefmark{4},
Renda Han \IEEEauthorrefmark{5},\\
Xiangxiang Lang \IEEEauthorrefmark{5}, 
Cuicui Luo \IEEEauthorrefmark{1},
}

\IEEEauthorblockA{\IEEEauthorrefmark{1} University of Chinese Academy of Science, Beijing, China \\ 
}
\IEEEauthorblockA{\IEEEauthorrefmark{2}Harbin Institute of Technology, Harbin, China
}
\IEEEauthorblockA{\IEEEauthorrefmark{3}City University of Hong Kong, Hong Kong SRA, China
}
\IEEEauthorblockA{\IEEEauthorrefmark{4}National University of Defense Technology, Changsha, China}
\IEEEauthorblockA{\IEEEauthorrefmark{5}Hainan University, Hainan, China}
\IEEEauthorblockA{\IEEEauthorrefmark{5}Corresponding Author}
}
\maketitle    
\begin{abstract}
Fraud detection is crucial in social service networks to maintain user trust and improve service network security. Existing spectral graph-based methods address this challenge by leveraging different graph filters to capture signals with different frequencies in service networks. However, most graph filter-based methods struggle with deriving clean and discriminative graph signals. On the one hand, they overlook the noise in the information propagation process, resulting in degradation of filtering ability. On the other hand, they fail to discriminate the frequency-specific characteristics of graph signals, leading to distortion of signals fusion. To address these issues, we develop a novel spectral graph network based on information bottleneck theory (SGNN-IB) for fraud detection in service networks. SGNN-IB splits the original graph into homophilic and heterophilic subgraphs to better capture the signals at different frequencies. For the first limitation, SGNN-IB applies information bottleneck theory to extract key characteristics of encoded representations.  For the second limitation, SGNN-IB introduces prototype learning to implement signal fusion, preserving the frequency-specific characteristics of signals. Extensive experiments on three real-world datasets demonstrate that SGNN-IB outperforms state-of-the-art fraud detection methods.
\end{abstract}
\section{Introduction}
\label{sec:Intro}
The rapid growth of digital service networks has transformed how services are delivered across industries, enabling seamless interactions across platforms, from financial services to e-commerce. However, this transformation has introduced new risks, particularly from sophisticated fraud schemes that undermine service quality, erode customer trust, and threaten operational stability. In service-oriented industries, where transaction networks and customer relationships form graph-structured systems, leveraging advanced analytics to address these risks is becoming a critical area for data-driven decision-making \cite{WOS:001128916400001}. This is particularly evident in financial platforms, where transaction records structured as graphs can reveal intricate patterns characteristic of fraudulent behavior. Developing effective fraud detection methods is essential, not only for enhancing system security but also for maintaining user trust and protecting the reputation of online platforms. As digital fraud schemes continue to grow in complexity, it is crucial to refine and advance graph-based detection methods to keep pace with emerging threats.

In this context, graph neural networks (GNNs) have emerged as a transformative technology for social service networks due to their exceptional ability to perceive interactive information, as demonstrated in various social service scenarios, such as fraud detection \cite{WOS:000423940800009, SEFraud}. GNNs are particularly well-suited for identifying risky and fraudulent behaviors that may be hidden within dense, high-dimensional interactive information. By integrating both interaction data and user-specific attributes, GNNs can detect suspicious activities with high accuracy, significantly enhancing the security of digital service platforms and establishing a more trustworthy online environment.

However, GNN-based fraud detection faces two main challenges: (1) \textbf{Data imbalance}. In real-world service ecosystems, fraudulent entities (such as fake accounts, malicious transactions, or service abuse) are often a minority within the network. The dominance of legitimate service nodes and regular interactions makes it difficult for detection models to capture the subtle anomalies associated with fraudulent behavior. This imbalance reduces the model's sensitivity to minority-class samples and weakens its ability to differentiate between normal service patterns and sophisticated fraud tactics, ultimately lowering both detection accuracy and generalization performance.

(2) \textbf{Heterophily}. Traditional GNNs, designed around homophily (the assumption that connected nodes exhibit similar features and behaviors), are poorly suited for service fraud detection. A significant limitation of these models is the over-smoothing effect, which is especially problematic in service networks. These models assume that interconnected nodes in a network share similar features and behaviors, thereby diminishing the ability to distinguish between linked entities. Fraudsters exploit this design flaw by creating cross-service-cluster relationships, such as generating high-frequency interactions or embedding themselves within legitimate transaction pathways to hide their fraudulent actions. Through these heterophilic strategies, fraudulent nodes can contaminate their local neighborhoods, obscuring their anomalous behavior and evading detection by GNNs. As a result, GNNs fail to identify the abnormal patterns, operational irregularities, and behavioral deviations that distinguish malicious users from legitimate participants in service networks.

To address these challenges, existing methods primarily focus on spatial domain analysis, which includes strategies like attention mechanisms \cite{LiuSAF0Y21}, resampling techniques \cite{H2-FDetector}, and auxiliary loss functions \cite{H2-FDetector}. For example, attention mechanisms can dynamically allocate the weights to the neighbors and manage to boost the contributions of nodes with high affinity; resampling techniques can adaptively determine which neighboring nodes to retain through feedback. However, these methods often face high computational costs and may alter the underlying structure of the service network. Recently, spectral domain analysis has been explored as a promising alternative \cite{IDGL, SplitGNN}. By filtering high- and low-frequency signals in the service network, spectral GNNs are better equipped to capture the distinct characteristics of anomalies, offering improved efficiency and accuracy over spatial approaches. 

Despite these advancements, spectral GNN-based fraud detection still have poor ability to capture clean and discriminative latent representations, which can be attributed to the following limitations: (1) Although graph filters can capture signals in different frequency domains, these filters still assume that information interaction between nodes in the network is effective behavior, ignoring noise variables introduced by malicious propagation and irrelevant behavior patterns. (2) Prevalent solution to heterophily is leverage different graph filters to capture the signals in different frequency. However, these signals from different graph filters lack the frequency-specific semantic discrimination, which makes the model hard to explicitly identify signals characteristics with different frequency domain, resulting in the distortion of the fused signals at the fusion node.

To address these issues, we propose a novel spectral graph network based on information bottleneck theory (SGNN-IB) for fraud detection. SGNN-IB first splits the original graph into homophilic and heterophilic subgraphs using a heterophily-aware classifier. It then applies multi-scale graph filters to capture both low- and high-frequency signals from the subgraphs and the original graph. For the first limitation, SGNN-IB incorporates information bottleneck theory \cite{IB} to enhance the encoding quality of graph filters with different frequency, alleviating the noise interference in the encoded node embeddings. For the second limitation, SGNN-IB employs prototype learning to boost the semantic discrepancy between high- and low-frequency signals, thereby helping model to identify diverse graph signals and fuse frequency-specific graph signals.

In summary, our contributions are as follows:
\begin{itemize}
\item We present a novel SGNN-IB model to derive clean and discriminative characteristics for fraud detection, which employs an edge classifier to split the original graph into homophilic and heterophilic subgraphs and then leverages Beta wavelet graph filters to capture critical characteristics of fraudsters.
\item We introduce an IB-based loss function to decrease the noise in different signals and utilize prototype learning to capture the frequency-specific characteristics and improve the signals integration.
\item Extensive experiments on widely used datasets demonstrate that our method significantly outperforms baseline approaches. Additionally, our ablation study validates the effectiveness of each component in the SGNN-IB framework.
\end{itemize}

\section{Related work}
\subsection{Graph-based fraud detection}
Graph-based methods for fraud detection in service networks leverage the inherent topological structure of service interactions to facilitate information propagation across individuals. A major challenge in fraud detection is data imbalance, as fraudsters often blend in with legitimate users, making their presence hard to detect. GNN-based fraud detection methods typically use various strategies to mitigate the impact of data imbalance and improve detection accuracy \cite{CARE-GNN, PC-GNN}. For instance, DIG-In-GNN \cite{DIG-IN-GNN} enhances the message-passing process with guidance information, while ASA-GNN \cite{ASA-GNN} adopts adaptive sampling strategies to filter out noisy nodes and propagate more representative information. Although these methods can effectively mitigate the issue of outliers in service networks, but the sampling strategies may disrupt the inherent structure of service interactions, leading to the loss of important information.

Another challenge is that fraudsters often hide by frequently interacting with benign users, leading to heterophily—where connected nodes exhibit different patterns. To tackle this, GraphConsis \cite{GraphConsis} employs semantic embedding, neighboring information, and adaptive attention to mitigate the effects of heterophily. $H^{2}$-FDetector \cite{H2-FDetector} adopts an opposite aggregation mechanism to learn discriminative representations for neighbors from different categories. GAGA \cite{GAGA} introduces a group-based strategy to mitigate the impact of high heterophily. Although these methods are effective, they suffer from significant computational complexity.

Recent studies have used graph filters to capture both low- and high-frequency signals. For example, AMNet \cite{AMNet} employs Bernstein polynomials to extract frequency-specific signals, and IDGL \cite{IDGL} applies dual-channel graph convolution filters to propagate multi-scale frequency information. Additionally, some research addresses the ``right-shift'' phenomenon caused by heterophily, using Beta wavelet transformations as spectral filters to capture important information  \cite{BWGNN, SplitGNN}.

Many filter-based methods rely on sophisticated graph filters to update node features, achieving success in identifying fraudsters in service networks. These methods often use classical graph filters, such as polynomial and wavelet transformations, to capture both low- and high-frequency information. Given the complexity of graph structures, some approaches apply filters at different levels or perspectives, such as global vs. local views, homophilic vs. heterophilic views, and relation-based views, to enhance model representation. Despite these advances, such methods are still limited to obtain representative characteristics of nodes and vulnerable to noise interference across different frequency domains.
\subsection{Graph neural networks}
GNNs are one of the most powerful deep learning methods for analyzing topological data, which can be categorized into spectral methods and spatial methods.

Spectral methods operate in a transductive manner, reasoning from observed samples instead of unknown samples. Several spectral GNNs, such as GCN \cite{GCN} and DGCF \cite{DGCF}, use Laplacian spectral transformations to capture topological embeddings and propagate information through convolution operations. Spectral methods are effective for capturing multi-scale frequency information. For example, OptBasisGNN \cite{OptBasisGNN} focuses on finding an optimal basis for spectral space, while PolyFormer \cite{PolyFormer} introduces a self-attention mechanism to enhance the utility of polynomial tokens. NFGNN \cite{NFGNN} introduces node-oriented spectral filtering with a generalized translated operator to improve the expression of local information.

Spatial methods focus on inference and generalization, enabling models to learn from training data and test on unseen datasets. Classical spatial GNNs, such as GAT \cite{GAT} and GraphSAGE \cite{GraphSAGE}, use a message-passing mechanism to disseminate information across the graph structure. Recent advancements in spatial GNNs include Edgeless-GNN \cite{Edgeless-GNN}, an unsupervised inductive framework for learning node representations without edges, and TULMGAT \cite{TULMGAT}, a multi-scale graph attention network designed for trajectory-user linking problems. H-GAT \cite{H-GAT} is a heterogeneous graph attention network that captures latent preferences of individuals based on interactions between heterogeneous instances. TL-GNN \cite{TL-GNN} addresses the local permutation invariance problem in GNNs to improve feature extraction quality.

However, most universal GNNs exhibit limitations in addressing fraud detection tasks due to two fundamental challenges. First, the inherent class imbalance in service fraud detection datasets significantly hinders traditional GNNs' ability to effectively learn and represent the distinctive patterns of fraudulent samples. Second, the presence of heterophilic relationship patterns in service networks contradicts the fundamental assumption of homophily in standard GNN architectures, thereby compromising their detection performance.
\section{Preliminaries}
In this section, we provide imperative definitions and detailed descriptions of the research problem.
\subsection{Definitions}
\textbf{Definition 1} ($Graph$): Let a graph $\mathcal{G} = (\mathcal{V}, \mathcal{E}, \mathcal{X}, \mathcal{A}, \mathcal{Y})$ denotes a  service network. $\mathcal{V} = \{v_{1}, v_{2}, \cdots,  v_{N}\}$ represents node set of graph $\mathcal{G}$, where $N$ is the number of nodes. $\mathcal{E}$ is the edge set of graph $\mathcal{G}$ and $e_{uv} \in \mathcal{E}$ denotes an edge from node $u$ to node $v$. $\mathcal{X} \in \mathbb{R}^{N \times D}$ indicates the feature matrix of $N$ nodes, where $D$ is the feature dimension. $\mathcal{A} \in \mathbb{R}^{N \times N}$ is the adjacency matrix of $\mathcal{G}$. If $e_{uv}\in \mathcal{E}$, $a_{uv} \in \mathcal{A}= 1$, otherwise $a_{uv} = 0$. $\mathcal{Y}\in \mathbb{R}^{N \times 1}$ denotes the label of all nodes, where $y_v \in \mathcal{Y} = 0$ if node $v$ is a benign sample and $y_v \in \mathcal{Y}= 1$ if node $v$ is a fraudster.

\textbf{Definition 2} ($Multi-relation graph$): If there are different relations between nodes in the graph, $\mathcal{G} = (\mathcal{V}, \mathcal{E}_{r} |^{R}_{r=1}, \mathcal{X}, \mathcal{A}_{r} |^{R}_{r=1}, \mathcal{Y})$ can be denoted as a multi-relation graph, where $R$ is the number of relation categories. For simplicity, a multi-relation graph can be identified as  $\mathcal{G} = (\mathcal{X}, \mathcal{A}_{r} |^{R}_{r=1}, \mathcal{Y})$.
\begin{figure*}
    \centering
    \includegraphics[width=0.98\linewidth]{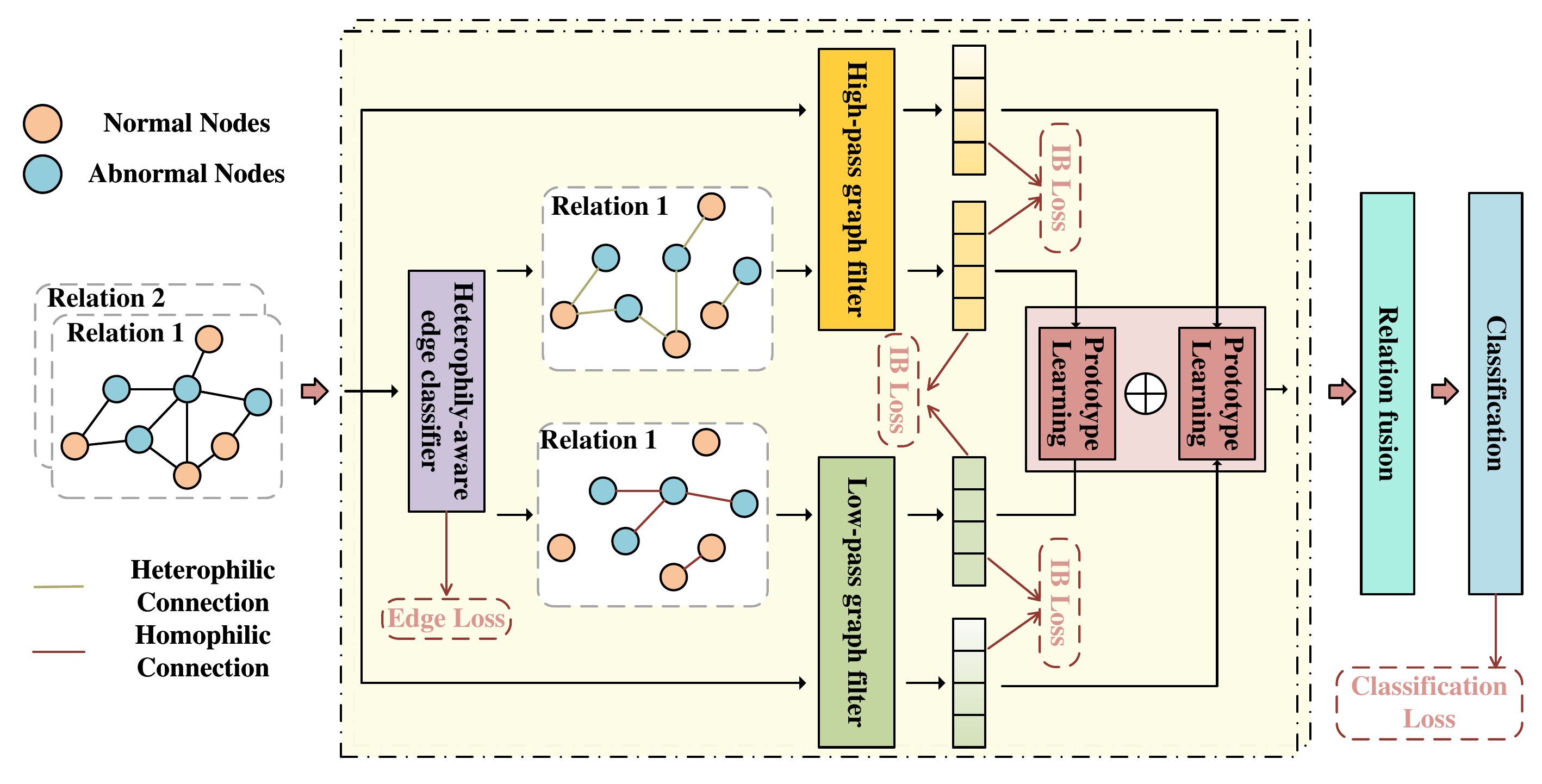}
    \caption{The framework of SGNN-IB. First, SGNN-IB leverages an edge classifier to perceive heterophilic subgraphs. Then, SGNN-IB utilizes multi-scale graph filters to obtain the high- and low-frequency signals in the graph. Subsequently, SGNN-IB integrate the signals from different frequency based on prototype learning. Finally, SGNN-IB is trained by the joint loss function, integrated with IB-loss.}
    \label{framework}
\end{figure*}
\subsection{Problem statement}
Given a multi-relation graph $\mathcal{G}$, the fraud detection problem in the service networks aims to discriminate fraud service providers and benign nodes in the graph $\mathcal{G}$ by exploiting a function $f_{\theta}(\cdot)$ with learnable parameters $\theta$ to project vector space of nodes into latent embedding space. The function is trained under a supervised learning paradigm, which uses node features $\mathcal{X}$, multi-relation adjacency matrices $\mathcal{A}_{r} |^{R}_{r=1}$ and node labels $\mathcal{Y}$ to optimize the function parameters $\theta$.

\section{Methodology}
In this section, we first introduce the overall SGNN-IB framework and then elaborate on the components of SGNN-IB and the training strategies.
\subsection{The SGNN-IB framework}
\begin{figure*}
    \centering
    \includegraphics[width=0.6\linewidth]{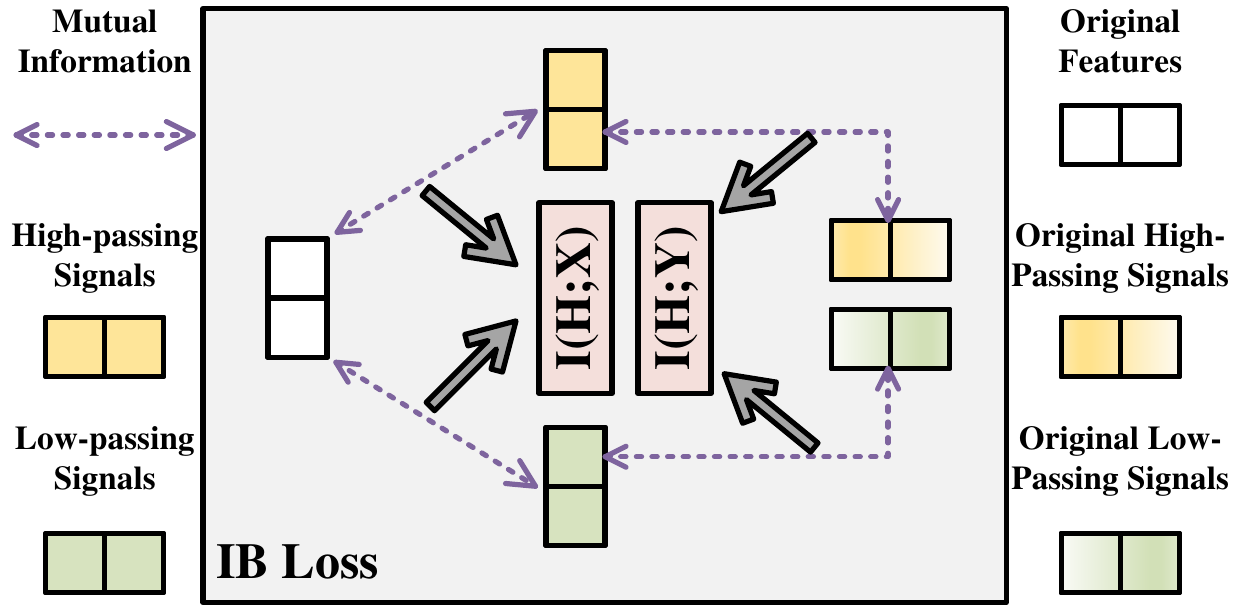}
    \caption{The architecture of IB loss. To solve the noise issue, the model leverages classical IB theory, maximizing the mutual information between the latent features and the ground truths and minimizing the mutual information between the latent features and the original features. Here, latent features denote the high-pass and low-pass signals and ground truths represent the band-pass signals. To solve the stochasticity issue, the model introduces the mutual information between high-pass and low-pass signals.}
    \label{IB LOSS}
\end{figure*}

The framework of SGNN-IB is shown in Figure \ref{framework}. First, SGNN-IB employs an edge classifier to identify and extract heterophilic subgraphs within the graph structure. Subsequently, SGNN-IB applies diverse graph filters to encode the original graph and specific subgraphs. the graph signals into high-pass, low-pass, and band-pass components, capturing diverse frequency-specific information. To enhance frequency-specific semantic discrimination, SGNN-IB introduces prototype learning to obtain the affinity of signals and performs information fusion. To conquer noise problem, SGNN-IB introduces an IB loss to alleviate the interference of noise in the process of information propagation. Finally,  SGNN-IB is trained with objective function which is comprised of IB loss, classification loss and edge loss, ensuring a balanced and comprehensive learning process. The architecture of IB loss is shown in Figure \ref{IB LOSS}.

\subsection{Heterophily-aware edge classifier}
Traditional GNNs are established on the assumption that the connections between nodes exhibit homophily, which means the connected nodes belong to the same category. In other words, traditional GNNs serve as smoothing function to the graph signals. However, many connections show heterophily, indicating that the connected nodes have different labels. Simply deploying traditional GNNs may dilute the categorical characteristics of nodes, which hinder the accurate node identification. Therefore, in order to avoid the loss of discriminative information in graph, it is important to split homophilic and heterophilic connections. 

To perceive the heterophily in graph topology, we design a heterophily-aware edge classifier, which aims to identify the edge type of each edge. In the context of training data containing labeled nodes, we meticulously establish homophilic and heterophilic edges based on the labels of source and target nodes in the training set. The edge classifier, designed as a binary classification model, leverages the feature representations of both the source node $u$ and the target node $v$ to predict the type of edge $e_{uv}$. This classifier is implemented using a multi-layer perceptron (MLP) architecture, thereby facilitating the discrimination between different edge types within the graph. 

For an edge $e_{uv}$ with the source node $u$ and target node $v$, the computations are as follows:
\begin{equation}
\label{edge0}
   \mathbf{h}_u = \sigma(\mathbf{W}_h \cdot \mathbf{x}_u + \mathbf{b}_h),
\end{equation}
\begin{equation}
\label{edge1}
   \mathbf{h}_v = \sigma(\mathbf{W}_h \cdot \mathbf{x}_v + \mathbf{b}_h),
\end{equation}
\begin{equation}
\label{edge2}
   \phi_{uv} = Sigmoid(\mathbf{W}[\mathbf{h}_u||\mathbf{h}_v||(\mathbf{h}_u - \mathbf{h}_v)]),
\end{equation}
\begin{equation}
\label{edge3}
   \pi_{uv} = 2*\phi_{uv} - 1,
\end{equation}
where $\sigma(\cdot)$ is a nonlinear activation function, $\mathbf{x}_u$ and $\mathbf{x}_v$ are respectively the original features of node $u$ and $v$, $\mathbf{W}_h$, $\mathbf{b}_h$ and $\mathbf{W}$ are learnable parameters of the feature transformation $\mathbf{h}_u$ and $\mathbf{h}_v$ are respectively transformed features of node $u$ and $v$, $Sigmoid(\cdot)$ is Sigmoid activation function, $[\cdot||\cdot]$ is concatenation function. $\pi_{uv}$ is limited to [-1, 1] to discriminate the heterophilic connections.

To partition the original graph into a homophilic subgraph $\mathcal{G}_{l}$ and a heterophilic subgraph $\mathcal{G}_{h}$, we leverage the prediction outcomes of all edges within the graph. The homophilic subgraph exclusively only comprises edges predicted to exhibit homophily, whereas the heterophilic subgraph merely encompasses edges anticipated to display heterophily. 

The precise classification of edges is of paramount importance for subsequent procedures, as it directly influences the quality of the resultant partitioned subgraphs. To this end, we devise an auxiliary loss function tailored for training the edge classifier. This loss is derived from the constructed training edge set $\mathcal{E}_{tr}$ and the corresponding prediction outcomes. The heterophily-aware edge classifier is optimized using the training edge set $\mathcal{E}_{tr}$ with the following loss function:
\begin{equation}
\label{edge3}
   \mathcal{L}_{H} = -\sum_{e_{uv} \in \mathcal{E}_{tr}} (y_{e_{uv}}\mathrm{log}(\phi_{uv})+(1-y_{e_{uv}})\mathrm{log}(1-\phi_{uv}),
\end{equation}
where $y_{e_{uv}}$ is the label of edge $e_{uv}$. If the edge exhibits homophily, the label $y_{e_{uv}}$ is 1, otherwise the label $y_{e_{uv}}$ is 0.
\subsection{The design of graph filter and information fusion}
Upon dividing the original graph, the resultant homophilic subgraph $\mathcal{G}_{homo}$ manifests an enrichment of low-frequency signals, whereas the heterophilic subgraph $\mathcal{G}_{heter}$ predominantly exhibits high-frequency signals. To capture signals within distinct frequency bands, diverse filters are applied to these partitioned graphs. Notably, since the splitting process yields frequency-specific signals from the original graph, subgraphs inevitably lose the holistic structural information contained within the original graph. To bolster the overall semantic richness and the fidelity to original information, it is imperative to also apply filters to the original graph.

Formally, consider the original graph $\mathcal{G}$, alongside the predicted homophilic subgraph $\mathcal{G}_{homo}$ characterized by its Laplacian $L_{homo}$, and the predicted heterophilic subgraph $\mathcal{G}_{heter}$ with $L_{heter}$. Given the model's need to discern signals of varying frequencies across these three graphs, a versatile band-pass filter becomes indispensable. Crafting an apt graph filter for the partitioned subgraphs presents a non-negligible challenge, as contemporary GNNs predominantly leverage low-pass filters \cite{SplitGNN}. Recently, research endeavors have introduced methodologies to learn arbitrary graph filters via polynomial approximation or Transformer architectures, exemplified by PolyFormer \cite{PolyFormer} and Specformer \cite{Specformer}. However, these methodologies fall short in the context of fraud detection tasks, where the minute proportion of fraudulent nodes within the graph exacerbates the issue of severe class imbalance. Consequently, high-frequency signals become relatively scant, leading the trained filter to potentially demonstrate a propensity for prioritizing low-frequency signals.

Consequently, we adopt design band-pass filters based on Beta wavelet \cite{BWGNN} to capture distinct frequency bands. Based on Beta distribution, Beta wavelet transformation is defined as follows:
\begin{equation}
\label{beta1}
   f(x;\alpha,\beta)=\frac{\Gamma(\alpha+\beta)}{\Gamma(\alpha)\Gamma(\beta)}x^{\alpha-1}(1-x)^{\beta-1} ,x \in [0, 1],
\end{equation}
where $\Gamma(\cdot)$ is Gamma distribution, $\alpha$ and $\beta$ is the parameters of Beta distribution. Given the eigenvalues of the normalized graph Laplacian $\mathbf{L} \in [0, 2]$, we leverage $f^{*}(x;\alpha,\beta)=\frac{1}{2}f(\frac{1}{2}x;\alpha,\beta)$ to cover the whole spectral range of $\mathbf{L}$. 

For simplicity, we constrain the $\alpha, \beta \in \mathbb{N}^{+}$ and only generate a low-pass filter $f^{*}_{low}(x;\alpha,\beta)$ and a high-pass filter $f^{*}_{high}(x;\beta,\alpha)$ to avoid computational complexity problem. 

Then we apply the high $f^{*}_{homo}$ to $\mathbf{L}_{homo}$ to capture low-frequency information from $\mathcal{G}_{homo}$. Correspondingly, we can obtain high-frequency signals by deploying $f^{*}_{heter}$ on the normalized Laplacian $\mathbf{L}_{heter}$ of $\mathcal{G}_{heter}$. The formulations can be defined as follows:
\begin{equation}
\label{low}
  \mathbf{H}_{low}=f^{*}_{low}(\mathbf{L}_{homo}, \mathbf{H})=f_{low}(\frac{1}{2}\mathbf{L}_{homo};\alpha,\beta)\mathbf{H},
\end{equation}
\begin{equation}
\label{high}
   \mathbf{H}_{high}=f^{*}_{high}(\mathbf{L}_{heter}, \mathbf{H})=f_{high}(\frac{1}{2}\mathbf{L}_{heter};\beta, \alpha)\mathbf{H},
\end{equation}
where $\mathbf{H}$ is the features matrix. Then, we integrate the obtained signals from different frequency domains:
\begin{equation}
\label{high-low}
  \hat{\mathbf{H}}=\Phi(\mathbf{H}_{high}, \mathbf{H}_{low}),
\end{equation}
where $\Phi(\cdot)$ is adaptive frequency fusion function, which is illustrated in Section \ref{proto}.

We have derived representations from both homophilic and heterophilic subgraphs utilizing low- and high-pass filters. Nevertheless, the structural integrity of these two subgraphs remains incomplete. To enhance the expressive power of our model, we employ the band-pass filters on the original graph and generate fused embeddings of band-pass filters.
\begin{equation*}
\label{band}
  \mathbf{H}^{o}_{i}=f^{*}_{i}(\mathbf{L}_{o}, \mathbf{H})=f_{i}(\frac{1}{2}\mathbf{L}_{o};\{\alpha,\beta\})\mathbf{H},
\end{equation*}
\begin{equation}
\label{band}
  \hat{\mathbf{H}}^{o}=\Phi(\mathbf{H}^{o}_{high}||\mathbf{H}^{o}_{low}),
\end{equation}
where $i \in\{high, low\}$, $\mathbf{L}_{o}$ is the normalized graph Laplacian of the original graph $\mathcal{G}$ and $H^{o}_{i}$ represents the transformed features by single low- or high-pass filter.

To protect the original semantic information of node features, the ultimate embedding of the node is constructed by concatenating the filtered representations and the linearly transformed residual representations from the original graph. 
\begin{equation}
\label{band}
  \Bar{\mathbf{H}}=\sigma(\mathbf{W}_{o}[\hat{\mathbf{H}}^{o}, \hat{\mathbf{H}}]).
\end{equation}

In practical scenarios, the majority of fraud graphs encompass a diverse relation. After acquiring representations for each relation, we integrate the node representations stemming from these various relations, thereby constructing the definitive embedding for the nodes. For the sake of brevity, we have omitted the explicit representation of these relations in the aforementioned equations. The relation fusion formulation can be defined as follows:
\begin{equation}
\label{relation}
  \mathbf{H}_{all} = \mathbf{W}_{r}||_{r=1}^{R}\Bar{\mathbf{H}}_{r},
\end{equation}
where $\Bar{\mathbf{H}}_{r}$ is the ultimate filtered embedding in homogeneous graph under relation $r$, $R$ is the relation set of graph $\mathcal{G}$ and $\mathbf{W}_{r}$ is the learnable weights.

\subsection{Frequency-specific feature fusion based on prototype learning}
\label{proto}
The high-frequency and low-frequency should reflect the behavior characteristics of nodes in different frequency domain modes. However,  due to the interactive pattern of nodes, these signals lack discernible frequency-specific semantic information, which loses significant discrimination after feature fusion. Therefore, we introduce an adaptive frequency fusion function $\Phi(\cdot, \cdot)$, a prototype learning mechanism, to enhance the semantic representations in each frequency domain.

Take high-frequency features as an example. Given the latent representations of frequency domain $\mathbf{H}_{high}$, we first calculate the prototype of high-frequency domain:
\begin{equation}
\label{proto1}
  \mathbf{c}_{high} = \mathrm{Readout}(\mathbf{H}_{high}),
\end{equation}
where $Readout(\cdot)$ is average readout function. 

Then we can obtain the affinity score of the node features with prototype:
\begin{equation}
\label{proto2}
  s_{high} = \frac{1}{\mathrm{len}(\mathbf{H})}\sum_{i=1}^{\mathrm{len}(\mathbf{H})} \mathrm{cos}(\mathbf{h}_{(i, high)}, \mathbf{c}_{high}),
\end{equation}
where $\mathbf{h}_{(i, high)}=\mathbf{H}[i, :]$, $\mathrm{cos}(\cdot, \cdot)$ denotes the Euclidean distance, and $\mathrm{len}(\mathbf{H})$ denotes the sample size in $\mathbf{H}$. Similarly, we can obtain the affinity score $s_{low}$ in low-frequency domain. A higher score indicates that the frequency-specific characteristics are more representative. 

To enhance the frequency-specific semantic discrimination, the fused representations should approach to frequency domain signals with high affinity. Therefore, we integrate the signals from high-frequency and low-frequency domain based on the affinity score:
\begin{equation}
\label{proto3}
  \Phi(\mathbf{H}_{high}, \mathbf{H}_{low}) = \frac{s_{high}}{s_{high}+s_{low}} \mathbf{H}_{high}+ \frac{s_{low}}{s_{high}+s_{low}}\mathbf{H}_{high}.
\end{equation}
To capture signals within distinct frequency bands, diverse filters are applied to these partitioned graphs. Notably, since the splitting process yields frequency-specific signals from the original graph and there is interference in the propagation of information in interactive behavior, subgraphs inevitably lose the holistic structural information contained within the original graph. To bolster the overall semantic richness and the fidelity to original information, it is imperative to also apply filters to the original graph.

\subsection{IB-based representation denoising}
Even though high-pass and low-pass filters encapsulate distinct semantic information within graphical representations, as illustrated in Section \ref{sec:Intro}, there is interference in the propagation of information in interactive behavior, which results in noise problems in the propagation of information. These issues leave the graph filtering capability constrained and hindering the generation of sufficiently discriminative representations across diverse frequency domains.

To this end, we introduce the IB theory to improve the quality of latent representations against noise. According to IB theory, the training objective is twofold: (1) to maximize mutual information between encoded embeddings $\mathbf{H}$ and labels $\mathbf{Y}$, and (2) to minimize mutual information between the encoded embeddings $\mathbf{H}$ and the node features $\mathbf{X}$.
\begin{equation}
\label{IB_theory}
  \mathop{argmin}_{\mathbf{H}} \quad -I(\mathbf{H};\mathbf{Y}) + \mu\cdot I(\mathbf{H};\mathbf{X}),
\end{equation}
where $\mu$ is a balanced coefficient. The IB Theory can compress the information within input data to distill and preserve the most task-relevant knowledge, effectively reducing noise and redundant information while extracting the most predictive and useful features.

Based on this idea, we develop an IB-based information-enhancing module to improve the quality of graph filters and provide more optimization guidance for signals in different frequencies. First, our basic objective function is consistent with IB theory: (1) to maximize the mutual information between the latent embeddings $\mathbf{H}$ and the labels $\mathbf{Y}$, and (2) to minimize the mutual information between the latent embeddings and input features $\mathbf{X}$. However, due to the lack of prior knowledge of different frequency signals, it is impractical to calculate the mutual information directly using ground truth labels. To this end, we regard the latent embeddings from the encoded original graph using different graph filter as the labels $\mathbf{Y}$ and the representations encoded from the heterophilic and homophilic using corresponding graph filters as the latent embeddings $\mathbf{H}$. Then, the IB-based loss function can be defined as follows:
\begin{equation}
\label{I(H;Y)}
  I(\mathbf{H};\mathbf{Y}) = I(\mathbf{H}_{high};\mathbf{H}^{o}_{high}) + I(\mathbf{H}_{low};\mathbf{H}^{o}_{low}),
\end{equation}
\begin{equation}
\label{I(H;X)}
  I(\mathbf{H};\mathbf{X}) = I(\mathbf{H}_{high};\mathbf{H}) + I(\mathbf{H}_{low};\mathbf{H}).
\end{equation}


The overall IB-based loss function is defined by averaging each term of mutual information:
\begin{equation}
\label{I(H;X)all}
  \mathcal{L}_{IB} = \frac{1}{2} \times [-I(\mathbf{H};\mathbf{Y}) +  \mu \cdot I(\mathbf{H};\mathbf{X})].
\end{equation}

Through the implementation of the information-enhancing module based on IB theory, the graph filters obtain explicit guidance to effectively counteract noise within features. This ensures that the encoded representations not only preserve the vital characteristics of the original features but also meticulously filter out redundant and irrelevant information. Additionally, the graph filters operate across different frequency channels, maintaining their specificity and ensuring that each channel remains relatively independent. This approach enables the generation of high-quality, fused features that are crucial for the accuracy of the model.

\subsection{Jointly training}
To distinguish between fraudulent and non-fraudulent nodes, we use the cross-entropy loss function for binary node classification. Specifically, given a training set $\mathcal{V}_{train}$, for each node $v \in \mathcal{V}_{train}$ with its corresponding label $y_v$, the classification loss function can be formulated as:
\begin{equation}
\label{cross loss}
  \mathcal{L}_{C}= -\sum_{v \in \mathcal{V}_{train}}[y_vlog(p_v) + (1 - y_v)log(1 - p_v)],
\end{equation}
\begin{equation}
\label{cross loss}
  p_v= softmax(\mathbf{h}_{all}).
\end{equation}

The joint loss function combines the loss from the node classification, the heterophily-aware edge classifier, and the information bottleneck (IB)-based enhancer, as follows:
\begin{equation}
\label{overall loss}
  \mathcal{L} = \mathcal{L}_{C} + \lambda\mathcal{L}_{H} + \eta \mathcal{L}_{IB},
\end{equation}
where $\lambda$ and $\eta$ are hyperparameters.

To address the class imbalance typical in fraud detection, we apply a sampling strategy during training. This involves selecting an equal number of benign nodes as fraudulent nodes for calculating the node classification loss. Similarly, to calculate edge classification loss, we ensure that an equal number of homophilic edges are considered compared to heterophilic edges.
\section{Experiments}
\subsection{Experimental setup}
\subsubsection{Datasets}
We execute experiments on three public fraud detection datasets, YelpChi \cite{CARE-GNN}, Amazon \cite{H2-FDetector} and FDCompCN \cite{SplitGNN}. In the YelpChi dataset, nodes represent reviews, and it includes three types of relations: 1) R-U-R represents the reviews posted by the same user, 2) R-S-R denotes reviews related to the same product with the same star rating, and 3) R-T-R stands for the reviews related to the same product posted in the same period. In the Amazon dataset, nodes represent users, with three types of relations: 1) U-P-U denotes users reviewing at least one same product, 2) U-S-U represents users having at least one same star rating within a specific period, and 3) U-V-U indicates subscribers with the top 5 percent mutual review text similarities. In the FDCompCN dataset,  nodes represent companies, and it includes three types of relations: 1) C-I-C represents companies that have investment relationships, 2) C-P-C indicates companies and their disclosed customers, and 3) C-S-C suggests companies and their disclosed suppliers. The dataset statistics are summarized in Table \ref{DATASETS}.
\begin{table*}[!htbp]
	\centering
	\caption{Statistics of datasets}
	\label{DATASETS}
    \resizebox{0.73\textwidth}{!}{
	\begin{tabular}{c c c c c c c} 
		\hline
		Dataset & Application & \#Node & Dimension & Fraud(\%)& Relation & \#Edge\\\hline
		\multirow{3}{*}{YelpChi} & \multirow{3}{*}{review} & \multirow{3}{*}{45954} & \multirow{3}{*}{32} & \multirow{3}{*}{14.53\%} & R-U-R & 98630\\
            ~ & ~ & ~ & ~ & ~ & R-T-R & 1147232\\
            ~ & ~ & ~ & ~ & ~ & R-S-R & 7693958\\
		\multirow{3}{*}{Amazon} & \multirow{3}{*}{review} & \multirow{3}{*}{11944} & \multirow{3}{*}{24} & \multirow{3}{*}{6.87\%} & U-P-U & 351216\\
            ~ & ~ & ~ & ~ & ~ & U-S-U & 7132958\\
            ~ & ~ & ~ & ~ & ~ & U-V-U & 2073474\\
		\multirow{3}{*}{FDCompCN} & \multirow{3}{*}{financial} & \multirow{3}{*}{5317} & \multirow{3}{*}{57} & \multirow{3}{*}{10.5\%} & C-I-C & 5686\\
            ~ & ~ & ~ & ~ & ~ & C-P-C & 760\\
            ~ & ~ & ~ & ~ & ~ & C-S-C & 1043\\\hline
	\end{tabular}}
\end{table*}

\subsubsection{Baselines}
We select ten baselines to validate the advancement of our model. We categorize the baselines into three groups: shallow methods, GNNs, and GNNs-based fraud detection frameworks. Among these, MLP and XGBoost are typically shallow methods base on feature learning, which ignore graph topology.
GCN, GAT ,FAGCN and GPR-GNN are GNNs-base methods. CARE-GNN, $H^{2}$-FDetector, BWGNN  and SEFraud are fraud detection framework based on GNNs.

\textbf{MLP} is a classical neural network basic architecture. 

\textbf{XGBoost} \cite{XGBoost} is tree structure-based is an optimized distributed gradient enhanced machine learning algorithm aimed at achieving high efficiency, flexibility, and portability

\textbf{GCN} \cite{GCN} is a classical transductive graph convolution neural network that leverages spectral transformation to extract graph information. 

\textbf{GAT} \cite{GAT} is an inductive graph attention network that adaptively propagates neighboring information according to learnable weights of neighbors. 

\textbf{FAGCN} \cite{FAGCN} is an improved GNN that adaptively learns low-frequency and high-frequency information in graph for more effective node representations.

\textbf{GPR-GNN} \cite{GPR-GNN} is a novel generalized PageRank GNN framework and jointly optimize node embeddings and structural information extraction according to dynamic GPR weights.

\textbf{CARE-GNN} \cite{CARE-GNN} is a camouflage-resistant graph neural network that utilizes reinforcement learning to achieve similarity-aware neighbor selections.

\textbf{$H^{2}$-FDetector} \cite{H2-FDetector} exploits homophilic and heterophilic connections in the graph by attention mechanism with opposite weights.

\textbf{BWGNN} \cite{BWGNN} leverages beta wavelet transformation to deal with energy distribution shifting in anomalies. 

\textbf{SEFraud} \cite{SEFraud} is an interpretable fraud detection framework that leverages learnable masks to capture expressive representations. 

\subsubsection{Evaluation settings}
Since the fraud detection problem exhibits data imbalance, we select four metrics to evaluate all models.
AUC is determined through the comparative ranking of prediction probabilities across all instances, thereby mitigating the impact of data imbalance on the detection process.
\begin{equation}
\label{AUC}
   AUC = \frac{1}{2}\sum_{i=1}^{m-1}(TPR_{i+1} - TPR)(FPR_{i+1}-FPR)
\end{equation}
where $TPR$ is the true positive rate, and $FPR$ is the false positive rate.

Recall evaluates the average, without weighting, of the fractions of actual fraudsters and actual normal users that a detector correctly identifies out of the total actual number of fraudsters and normal users, respectively.
\begin{equation}
\label{Recall}
   Recall = \frac{TP}{TP + FN}
\end{equation}
where $TP$ and $FN$ are true positive and false negative, respectively.

GMean integrates the recall and accuracy metrics of a classifier to assess the effectiveness of a classification model when dealing with an unbalanced dataset, which is defined as follows:
\begin{equation}
\label{GMean}
   GMean = \sqrt{TPR*FPR}
\end{equation}

F1-score is an evaluation metric that comprehensively considers the performance of both precision and recall.
\begin{equation}
\label{F1}
    F1-score = \frac{2 \times precision \times Recall}{precision + Recall}
\end{equation}
\subsubsection{Implementation details}
The experiments utilize PyTorch in Python 3.9.12, deploying a single NVIDIA A40 GPU, 40GB of RAM, and a 2.60GHz Xeon (R) Gold 6240 CPU. All the baselines can be reproduced by public source codes and Python dependencies.
\begin{table*}[!htbp]
	\centering
	\caption{Performance of the proposed SGNN-IB model and comparative model on three datasets. All results are in \%.}
	\label{Overall Performance}
  \resizebox{0.98\textwidth}{!}{
	\begin{tabular}{c c| c c c c | c c c c| c c c c}
		\hline
		\multirow{2}{*}{Method} & Dataset & \multicolumn{4}{c}{YelpChi} & \multicolumn{4}{c}{Amazon}& \multicolumn{4}{c}{FDCompCN}\\
		\cline{2-14}
		~& Metric & Recall & F1-macro & AUC & GMean & Recall & F1-macro & AUC & GMean & Recall & F1-macro & AUC & GMean\\\hline
		\multirow{8}{*}{Baselines} & XGBoost & 19.15 & 61.72 & 59.01 & 43.51 & 69.09 & 72.68 & 79.54 & 78.87 & 61.25 & 61.17 & 50.64 & 58.04\\
		~ & MLP & 69.37 & 61.48 & 77.43 & 70.73 & 78.18 & 72.95 & 87.78 & 82.93& 57.08 & 54.80 & 43.06 & 58.48\\
		\cline{2-14}
            ~ & GCN & 77.53 & 36.67 & 59.33 & 49.46 & 80.00 & 56.43 & 84.61 & 73.72& 52.92 & 51.01 & 40.89 & 43.95\\
            ~ & GAT & 62.15 & 42.77 & 56.13 & 53.13 & 80.00 & 71.46 & 88.03 & 83.04& 52.55 & 51.36 & 38.20 & 42.94\\
		~ & GPRGNN & 75.16 & 57.34 & 77.12 & 69.84 & 80.09 & 64.15 & 89.08 & 82.32& 56.40 & 47.52 & 50.31 & 52.09\\
		~ & FAGCN & 70.64 & 61.11 & 77.90 & 70.86 & 81.21 & 69.30 & 90.48 & 84.33& 57.90 & 48.48 & 51.59 & 49.50\\
		\cline{2-14}
		~ & CARE-GNN & 72.32 & 60.40 & 77.41 & 70.86 & 75.76 & 70.45 & 86.19 & 81.71& 57.21 & 43.59 & 49.00 & 50.10\\
            ~ & $H^{2}$-FDetector & $\underline{84.61}$ & 70.78 & 88.90 & 81.64 & 82.12 & $\underline{71.36}$ & 89.84 & 84.29& 55.96 & 48.33 & 47.89 & 49.10\\
            ~ & BWGNN & 82.56 & 72.32 & $\underline{89.72}$ & 81.92 & 83.94 & 69.43 & $\underline{91.91}$ & 84.67& $\underline{58.01}$ & 47.91 & 49.79 & 52.33\\
            ~ & SEFraud & 78.64 & $\underline{72.51}$ & 86.77 & $\underline{82.44}$ & $\underline{88.67}$ & 71.28 & 91.5 & $\underline{85.13}$ & 57.49 & $\underline{50.31}$ & $\underline{50.41}$ & $\underline{53.74}$\\
		Ours & SGNN-IB & $\mathbf{86.37}$ & $\mathbf{74.64}$ & $\mathbf{92.06}$ & $\mathbf{84.40}$ & $\mathbf{90.30}$ & $\mathbf{71.56}$ & $\mathbf{93.03}$ & $\mathbf{86.65}$ & $\mathbf{58.93}$ & $\mathbf{52.22}$ & $\mathbf{56.43}$ & $\mathbf{54.17}$ \\\hline
	\end{tabular}}
\end{table*}
\begin{table*}[!htbp]
	\centering
	\caption{Performance of the ablation experiments on three datasets. All results are in \%.}
	\label{ablation Performance}
  \resizebox{0.98\textwidth}{!}{
	\begin{tabular}{c| c c c c | c c c c| c c c c}
		\hline
		\multirow{2}{*}{Variants} & \multicolumn{4}{c}{YelpChi} & \multicolumn{4}{c}{Amazon}& \multicolumn{4}{c}{FDCompCN}\\
		\cline{2-13}
		~ & Recall & F1-macro & AUC & GMean & Recall & F1-macro & AUC & GMean & Recall & F1-macro & AUC & GMean\\\hline
            
		SGNN-IB$_{edge}$ & 84.32 & 68.28 & 85.62 & 74.36 & 83.21 & 66.34 & 84.51 & 81.51 & 51.21 & 46.32 & 50.11 & 48.29\\
		SGNN-IB$_{low}$ & $\underline{85.31}$ & 67.48 & 85.43 & 76.73 & 82.37 & 64.74 & 85.97 & 76.94 & 52.14 & 45.80 & 49.39 & 48.22\\
            SGNN-IB$_{high}$ & 83.53 & 66.67 & 83.33 & $\underline{79.46}$ & 80.44 & 63.48 & 82.11 & 79.55 & 51.39 & 43.82 & 50.73 & 46.77\\
		SGNN-IB$_{rel}$ & 82.64 & 66.15 & 77.90 & 74.86 & 83.64 & 65.54 & 88.54 & 80.28 & $\underline{55.83}$ & 49.33 & 50.81 & 48.96\\
            SGNN-IB$_{IB}$ & 81.53 & $\underline{67.57}$ & $\underline{89.13}$ & 79.26 & $\underline{88.99}$ & $\underline{69.45}$ & $\underline{90.42}$ & $\underline{84.68}$ & 53.93 & $\underline{50.48}$ & $\underline{53.61}$ & $\underline{51.28}$\\\hline
		SGNN-IB & $\mathbf{86.37}$ & $\mathbf{74.64}$ & $\mathbf{92.06}$ & $\mathbf{84.40}$ & $\mathbf{90.30}$ & $\mathbf{71.56}$ & $\mathbf{93.03}$ & $\mathbf{86.65}$ & $\mathbf{58.93}$ & $\mathbf{52.22}$ & $\mathbf{56.43}$ & $\mathbf{54.17}$ \\\hline
	\end{tabular}}
\end{table*}
\subsection{Overall performance}

The experimental results of our study are summarized in Table \ref{Overall Performance}. The best results are highlighted in bold, while the second-best results are underlined.

In general, graph-based approaches outperform feature-based shallow methods, as they are better equipped to capture the complex relational and interactive information embedded within the graph structure. Among the graph-based models, GNN-based fraud detection methods outperform traditional GNN models. This advantage stems from the ability of GNNs to handle the data imbalance commonly seen in fraud detection tasks, which often causes classical GNNs to suffer from the over-smoothing problem, especially when distinguishing fraudsters from benign users.

Looking specifically at shallow methods, we find that MLP performs better than XGBoost. This is because MLPs can adaptively learn and represent nonlinear relationships in the data, which enhances their performance in fraud detection tasks. Among GNN-based models, GPRGNN and FAGCN demonstrate better performance compared to traditional GCN and GAT models. GPRGNN excels by capturing both structural and feature information, using adaptive weights to filter out noisy nodes. FAGCN, on the other hand, is particularly effective at identifying fraud-related features by distinguishing between high- and low-frequency information in the graph.
Within the realm of GNN-based fraud detection, several advanced models have also shown promising results. CARE-GNN mitigates the challenge of data imbalance through sampling strategies.$H^{2}$-FDetector is skilled at detecting differences between homophilic and heterophilic pairs in the graph, which helps identify fraudsters more accurately. BWGNN addresses the issue of fraudsters' camouflage by combining graph filters and convolutional networks. SEFraud incorporates a trainable masking strategy to improve the interpretability of node features.

The proposed SGNN-IB outperforms all these baseline models. In comparison to the best performance in baselines, SGNN-IB shows an absolute improvement of 1.76\%, 2.13\%, 2.34\%, and 1.96\% in Recall, F1-Macro, AUC, and GMean on the YelpChi dataset.  On the Amazon dataset, SGNN-IB achieves absolute improvements of 1.63\%, 0.20\%, 1.12\%, and 1.52\%, respectively. For the FDCompCN dataset, SGNN-IB improves by 0.92\%,1.91\%, 6.02\%, and 0.43\% in Recall, F1-Macro, AUC, and GMean.

The success of SGNN-IB can be attributed to several key factors. First, SGNN-IB uses both low-pass and high-pass filters to selectively extract relevant information from homogeneous and heterogeneous structures, respectively. It also employs a prototype learning method to maintain the discriminative information of different frequency domain. In addition, to enhance the robustness of the filtering process against noise, SGNN-IB integrates an IB-based enhancement module. This module guides the graph filter, enabling it to generate high-quality, encoded features that improve fraud detection performance.

\subsection{Ablation experiments}
To evaluate the contribution of each component in the SGNN-IB framework, we conduct ablation studies by examining five variants: SGNN-IB without heterophily-aware edge classifier SGNN-IB$_{edge}$, SGNN-IB without low-pass filter SGNN-IB$_{low}$, SGNN-IB without high-pass filter SGNN-IB$_{high}$, SGNN-IB without relation fusion SGNN-IB$_{rel}$ and SGNN-IB without IB-based information enhancer SGNN-IB$_{IB}$. The results of these ablation experiments are presented in Table \ref{ablation Performance}, with the best results highlighted in bold and the second-best results underlined.

The results indicate that SGNN-IB outperforms all its variants, demonstrating the effectiveness of each component in the framework. Meanwhile, SGNN-IB$_{low}$ performs relatively close to SGNN-IB, while SGNN-IB$_{high}$ shows lower performance. This suggests that high-pass signals play a particularly important role in detecting fraudulent activities. Additionally, the performance of SGNN-IB$_{IB}$ reinforces the effectiveness of the IB-based information enhancement module, which contributes to noise reduction and improved model robustness.
\subsection{Sensitivity experiments}
\begin{figure*}[!htb]
\centering
\resizebox{0.98\textwidth}{!}{
\subfloat{
		\includegraphics[scale=0.27]{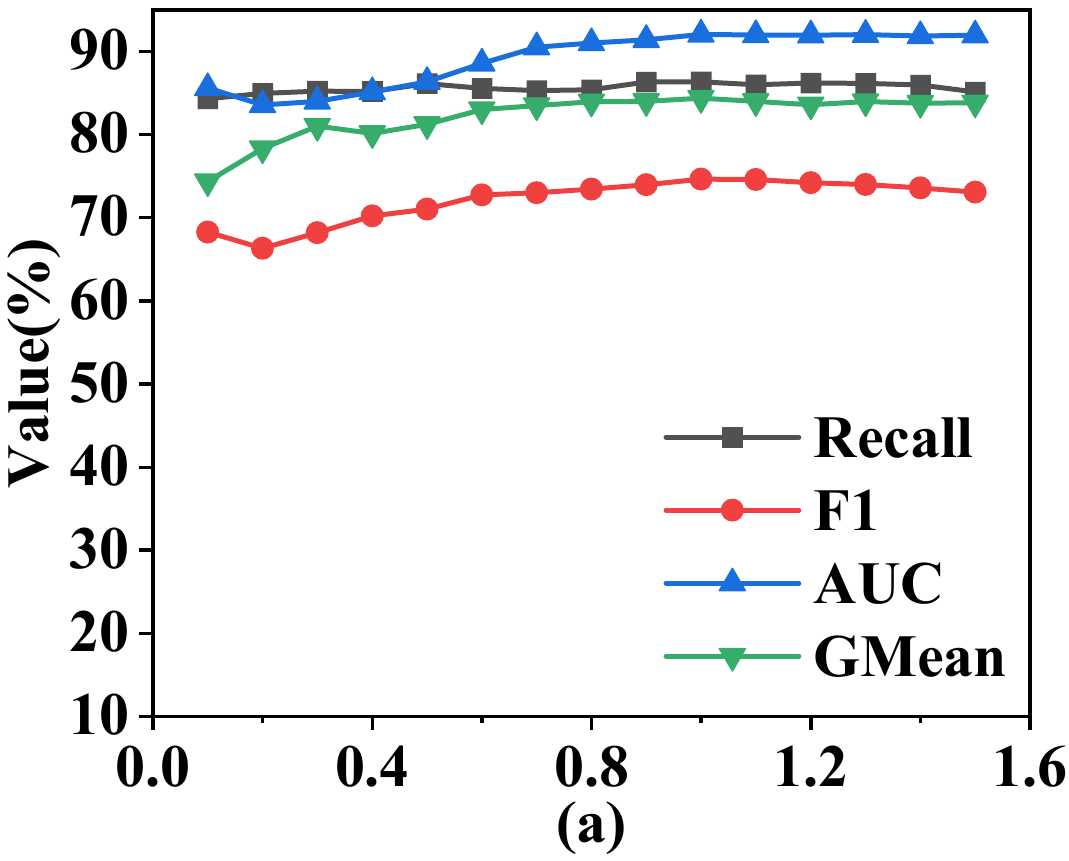} \label{yelplambda}}
\subfloat{
		\includegraphics[scale=0.27]{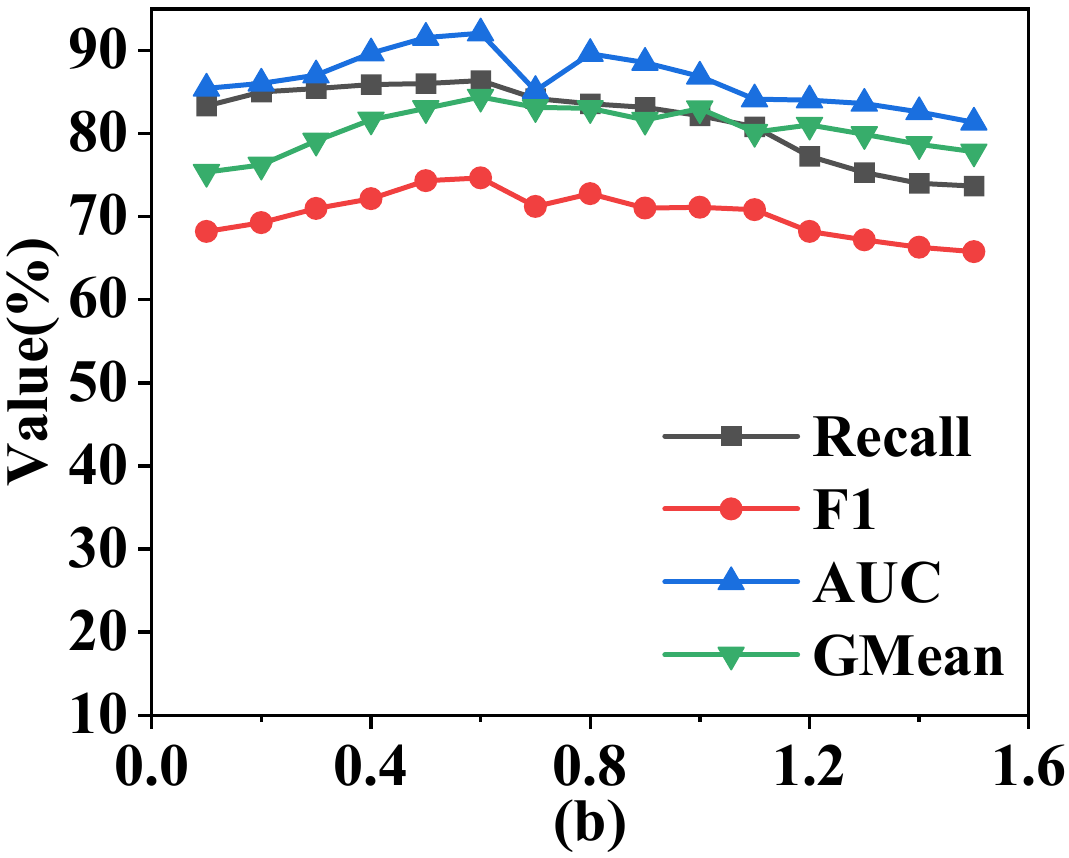} \label{yelpeta}}
  \subfloat{
		\includegraphics[scale=0.27]{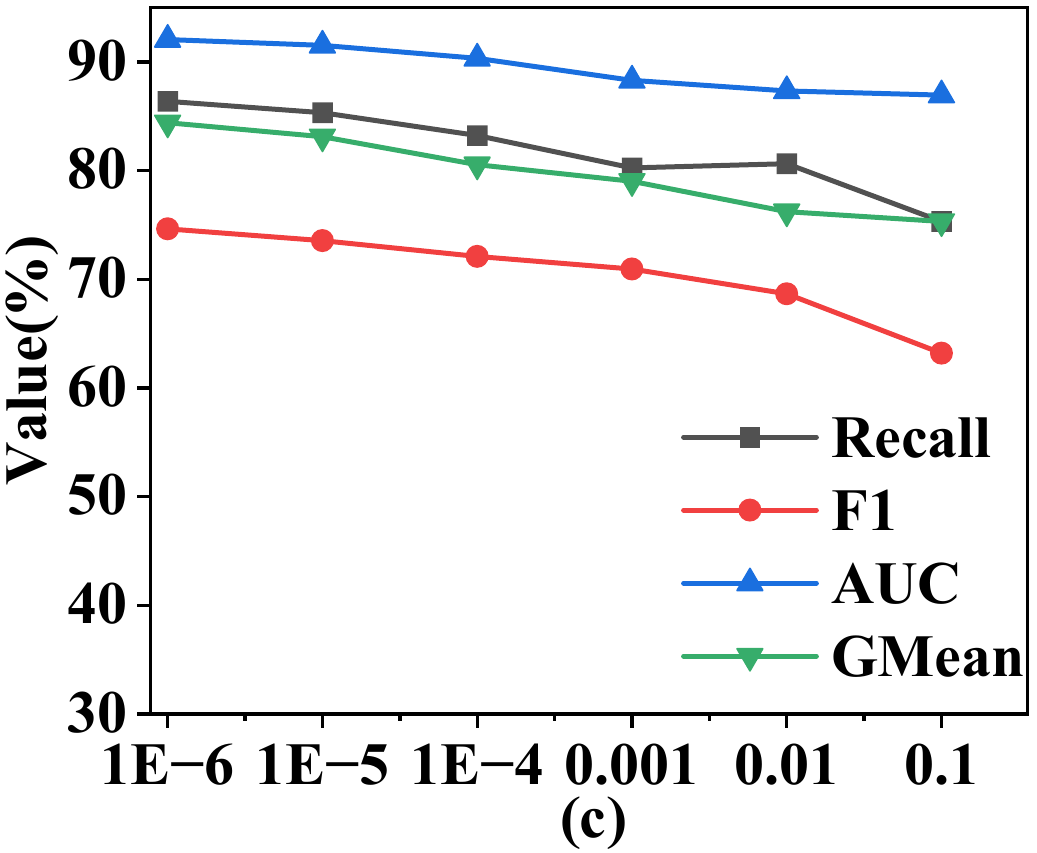} \label{yelpmu}}}
\caption{Sensitivity experimental results on YelpChi dataset: (a) Sensitivity results for parameter $\lambda$; (b) Sensitivity results for parameter $\eta$; (c) Sensitivity results for parameter $\mu$.}
\label{YelpChisensitivity}
\end{figure*}

\begin{figure*}[!htb]
\centering
\resizebox{0.98\textwidth}{!}{
\subfloat{
		\includegraphics[scale=0.27]{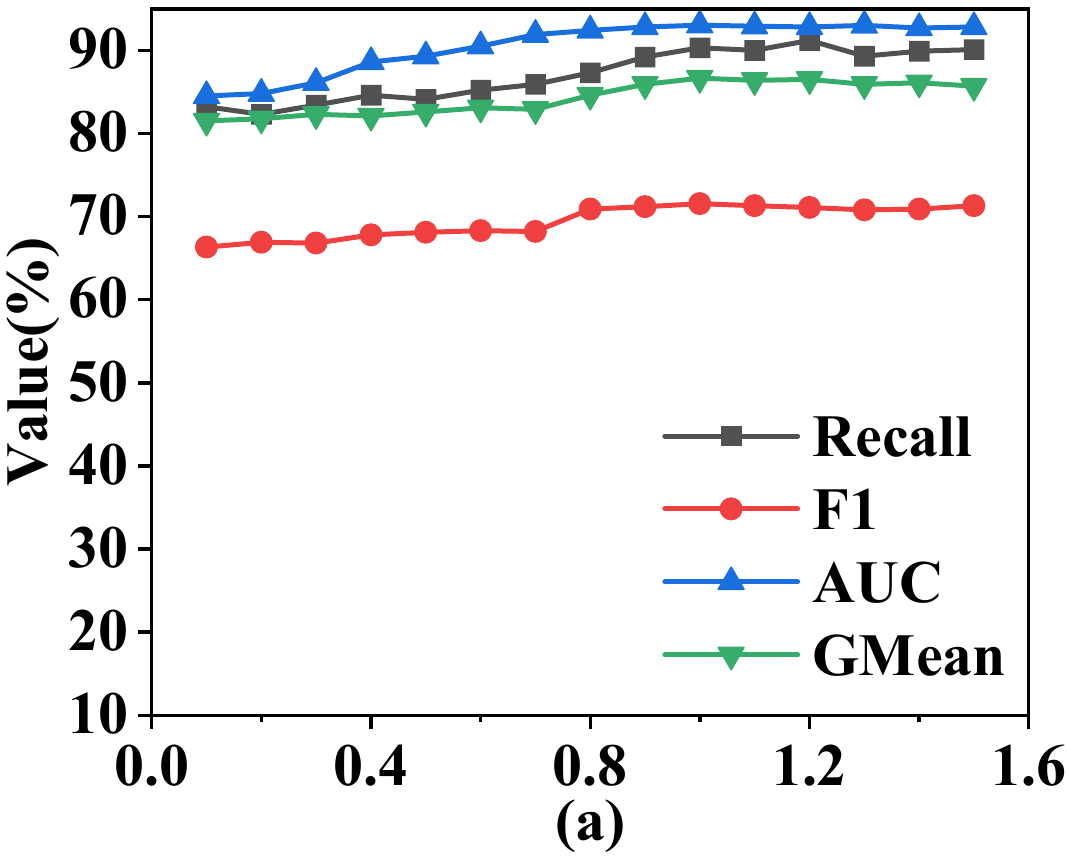} \label{amazonlambda}}
\subfloat{
		\includegraphics[scale=0.27]{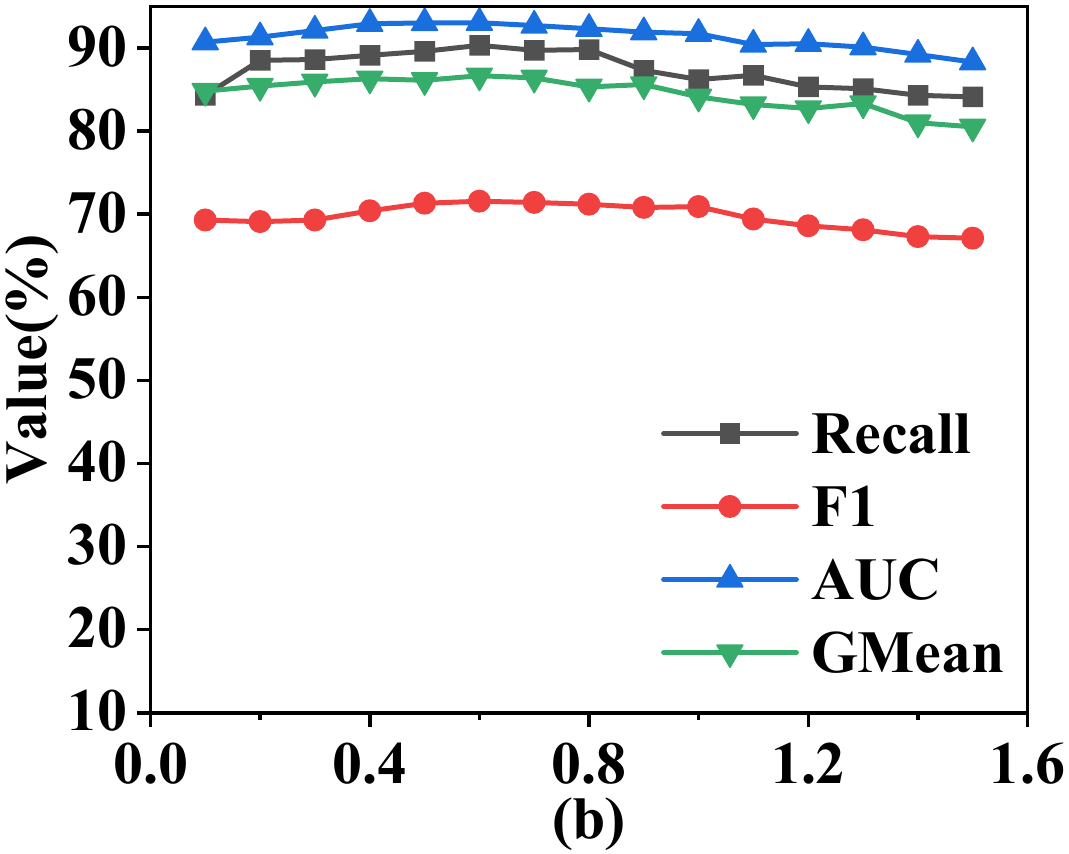} \label{amazoneta}}
  \subfloat{
		\includegraphics[scale=0.27]{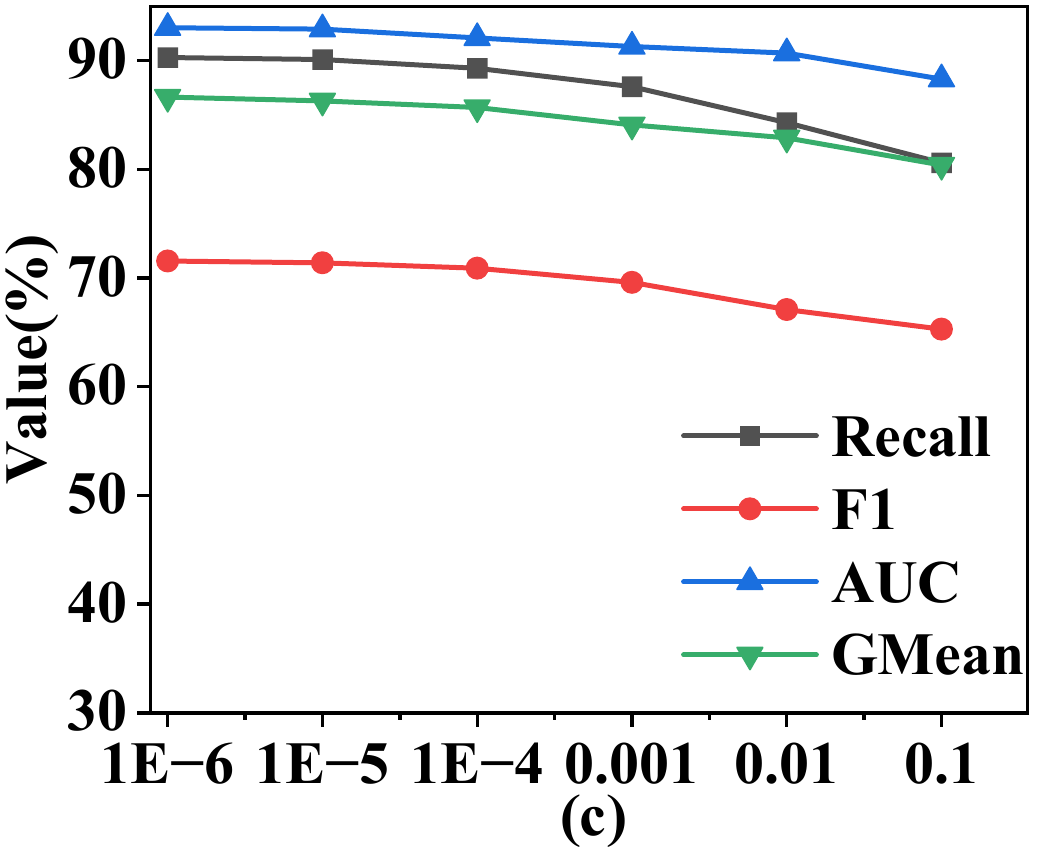} \label{amazonmu}}}
\caption{Sensitivity experimental results on the Amazon dataset: (a) Sensitivity results for parameter $\lambda$; (b) Sensitivity results for parameter $\eta$; (c) Sensitivity results for parameter $\mu$.}
\label{Amazonsensitivity}
\end{figure*}

\begin{figure*}[!htb]
\centering
\resizebox{0.98\textwidth}{!}{
\subfloat{
		\includegraphics[scale=0.25]{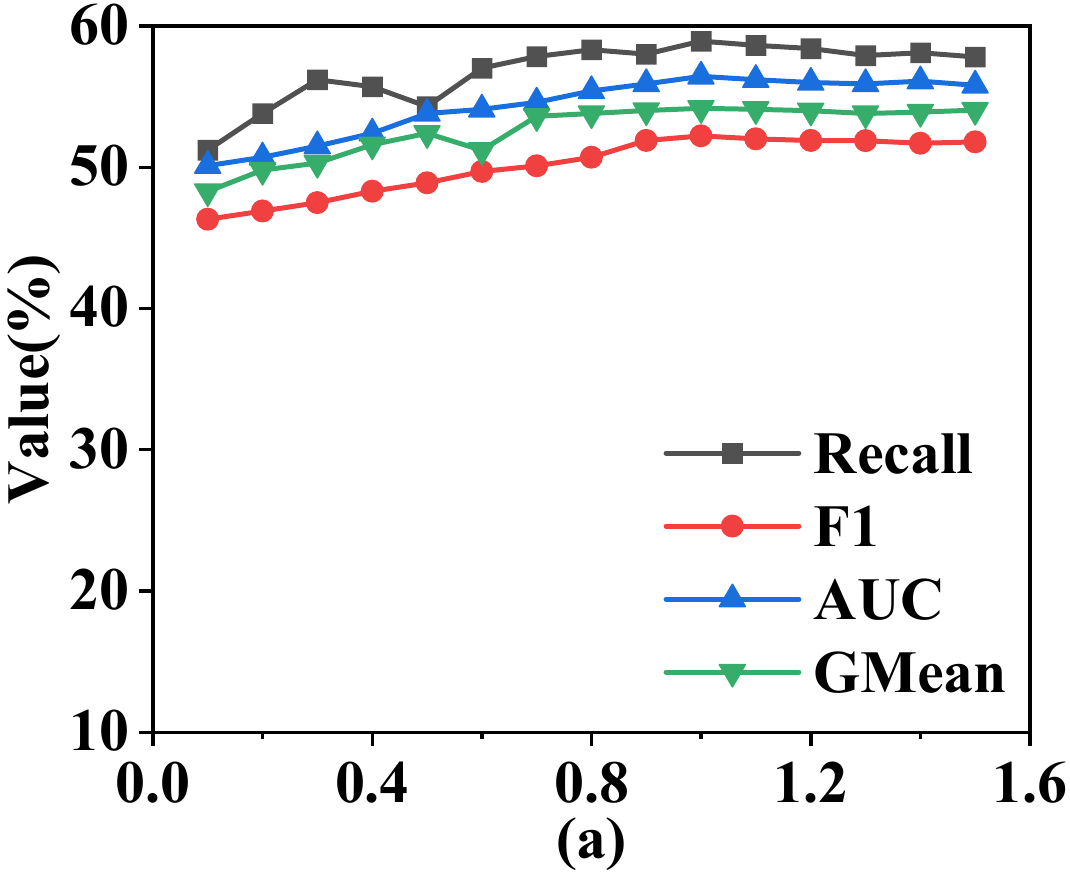} \label{FDlambda}}
\subfloat{
		\includegraphics[scale=0.25]{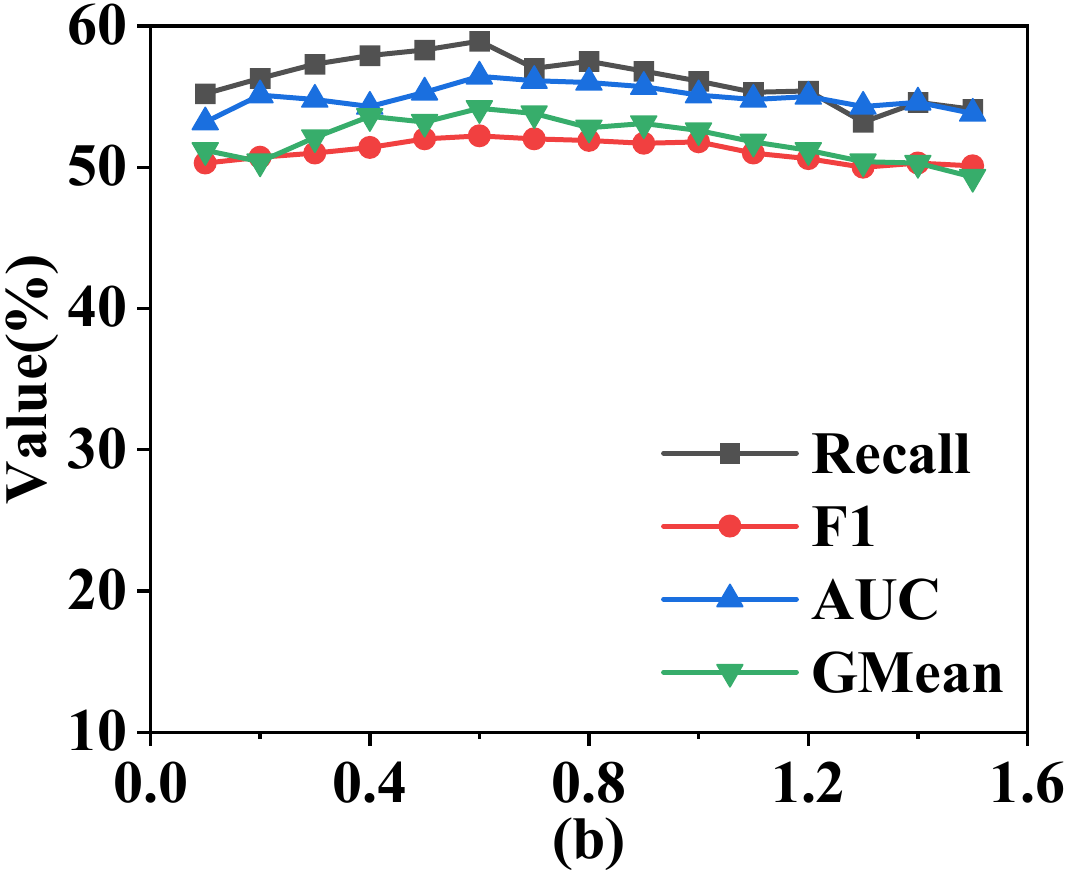} \label{FDeta}}
  \subfloat{
		\includegraphics[scale=0.25]{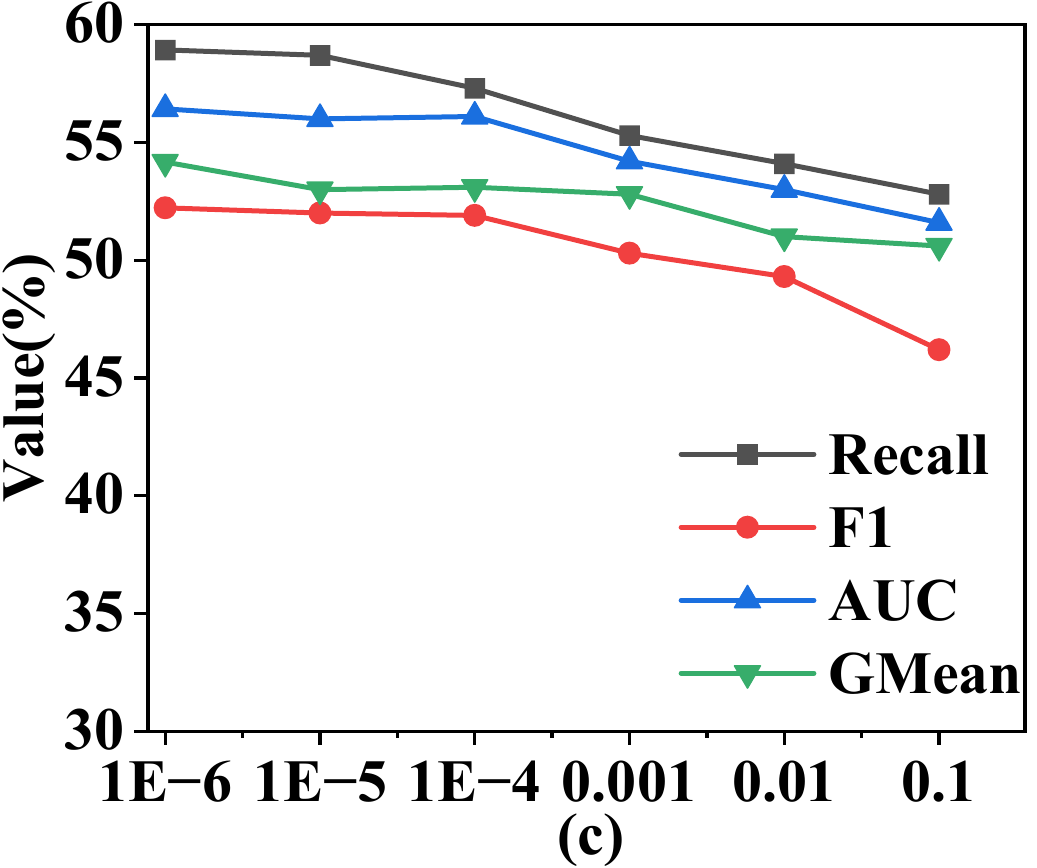} \label{FDmu}}}
\caption{Sensitivity experimental results on FDCompCN dataset: (a) Sensitivity results for parameter $\lambda$; (b) Sensitivity results for parameter $\eta$; (c) Sensitivity results for parameter $\mu$.}
\label{FDsensitivity}
\end{figure*}

We conduct sensitivity experiments by selecting three key model hyperparameters: $\mu$, $\lambda$, and $\eta$. The parameter $\mu$ controls the contribution of mutual information between the input features and the filtered features, as well as between different filter channels. The parameter $\lambda$ controls the influence of the heterophily-aware edge classifier, while $\eta$ controls the contribution of the information enhancement loss based on the information bottleneck (IB) theory. The values of $\lambda$ and $\eta$ range from 0.1 to 1.5, with a step size of 0.1. The range for $\mu$ is from 0.000001 to 0.1, with an exponential step size. The results of these sensitivity experiments for the YelpChi, Amazon, and FDCompCN datasets are shown in Figure \ref{YelpChisensitivity}, \ref{Amazonsensitivity}, and \ref{FDsensitivity}, respectively.

From the sensitivity experiments across these three datasets, we observe that the parameters $\lambda$ and $\mu$ have a significant impact on model performance, while $\eta$ plays a relatively minor role. Specifically, Take the YelpChi dataset as an example. As shown in Figure \ref{yelplambda}, a small value of $\lambda$ limits the effectiveness of the edge classifier, leading to incorrect identification of heterophilic edges. This misclassification hampers the capture of high-frequency signals, which are crucial for identifying fraudulent behavior, thus reducing the model's ability to detect fraudsters. On the other hand, increasing $\lambda$ enhances the classifier’s capacity, but its effect on performance is relatively small beyond a certain threshold. Figures \ref{yelpeta} and \ref{yelpmu} further show that $\eta$ and $\mu$ mainly affect the model’s ability to filter noise and extract key features. However, when these values are too large, the loss function tends to converge rapidly to negative values during training, resulting in a slight decline in performance. In particular, for $\mu$, which regulates the data purification and compression between the input data and the filtered features, smaller exponential values are more effective. This allows SGNN-IB to focus on the most essential components of the original features, improving its ability to capture the key information related to fraud.

Based on these findings, we determine the optimal settings for each dataset. For the YelpChi dataset, the best values for $\lambda$, $\eta$, and $\mu$ are 1, 0.6, and 0.000001, respectively, as shown in Figure \ref{YelpChisensitivity}. For the Amazon dataset, the optimal values are 1, 0.5, and 0.000001, as seen in Figure \ref{Amazonsensitivity}. Finally, for the FDCompCN dataset, the best choices for $\lambda$, $\eta$, and $\mu$ are 1, 0.6, and 0.000001, respectively, according to Figure \ref{FDsensitivity}.

\begin{figure*}[!htb]
\centering
\subfloat{
		\includegraphics[scale=0.27]{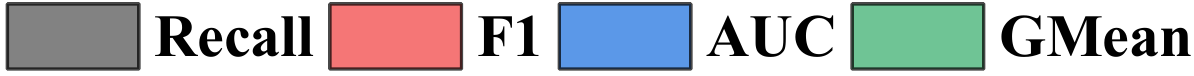} \label{tuli1}}\\
\resizebox{0.98\textwidth}{!}{
\subfloat{
		\includegraphics[scale=0.27]{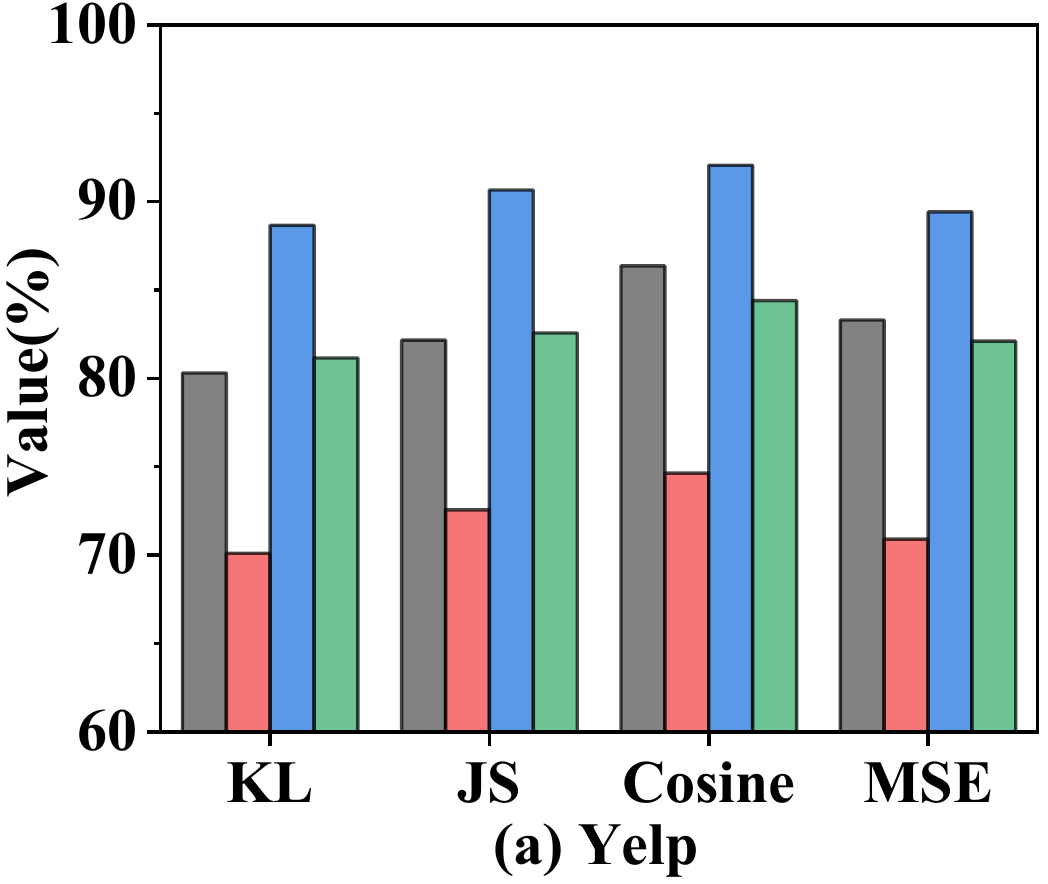} \label{yelpmetric}}
\hspace{-2mm}
\subfloat{
		\includegraphics[scale=0.27]{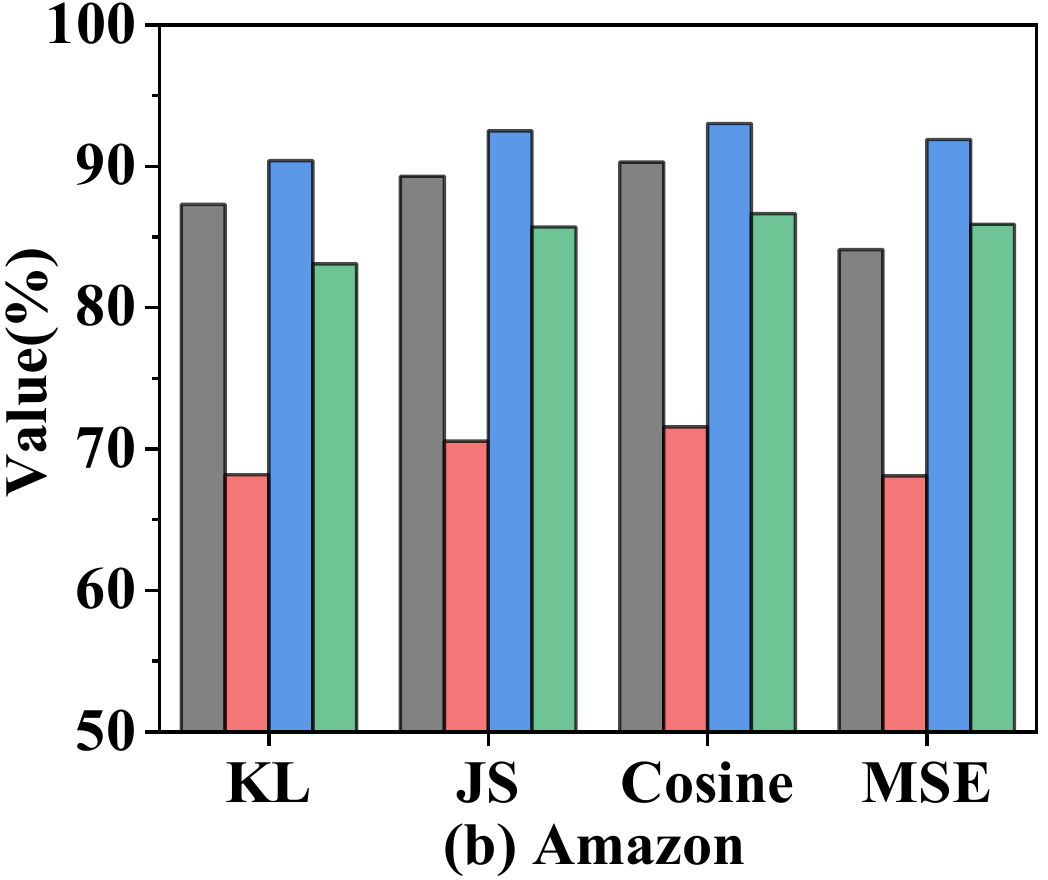} \label{amazonmetric}}
\hspace{-2mm}
\subfloat{
		\includegraphics[scale=0.27]{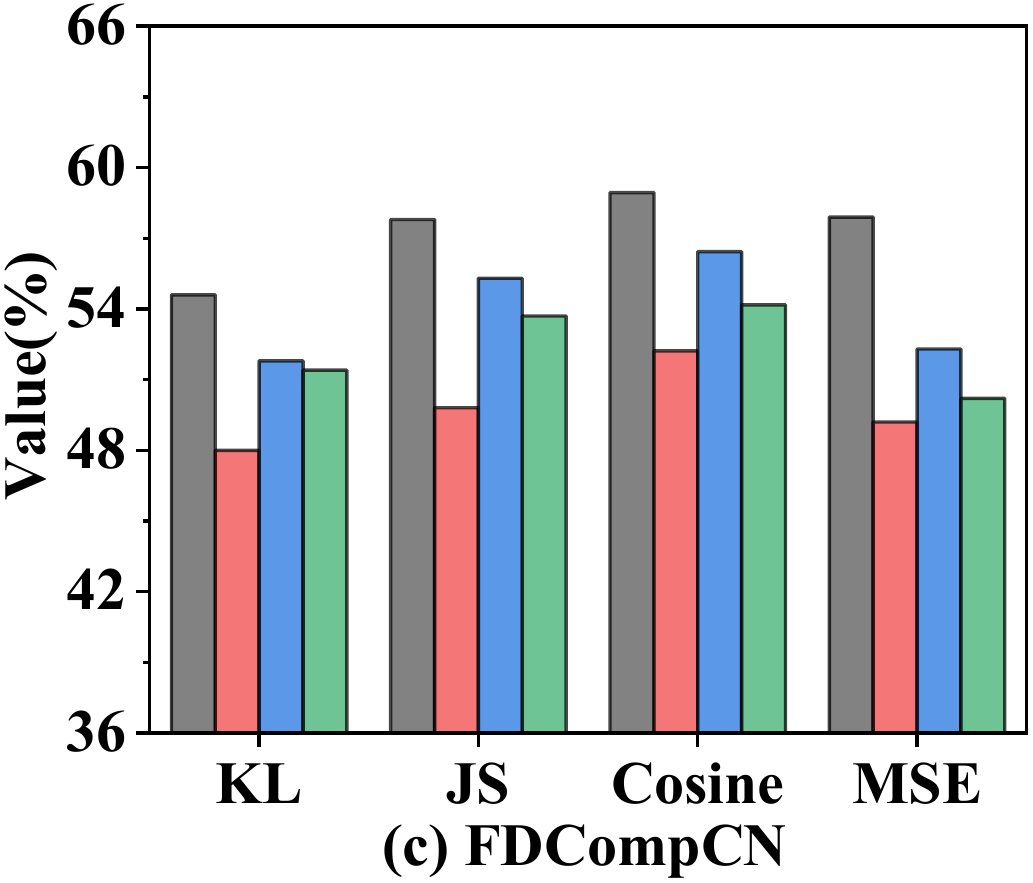} \label{FDmetric}}}
\caption{The performance of using different metrics for mutual information computation.}
\label{Model analysis1}
\end{figure*}

\begin{figure*}[!htb]
\centering
  \resizebox{0.95\textwidth}{!}{
\subfloat[YelpChi]{
		\includegraphics[scale=0.22]{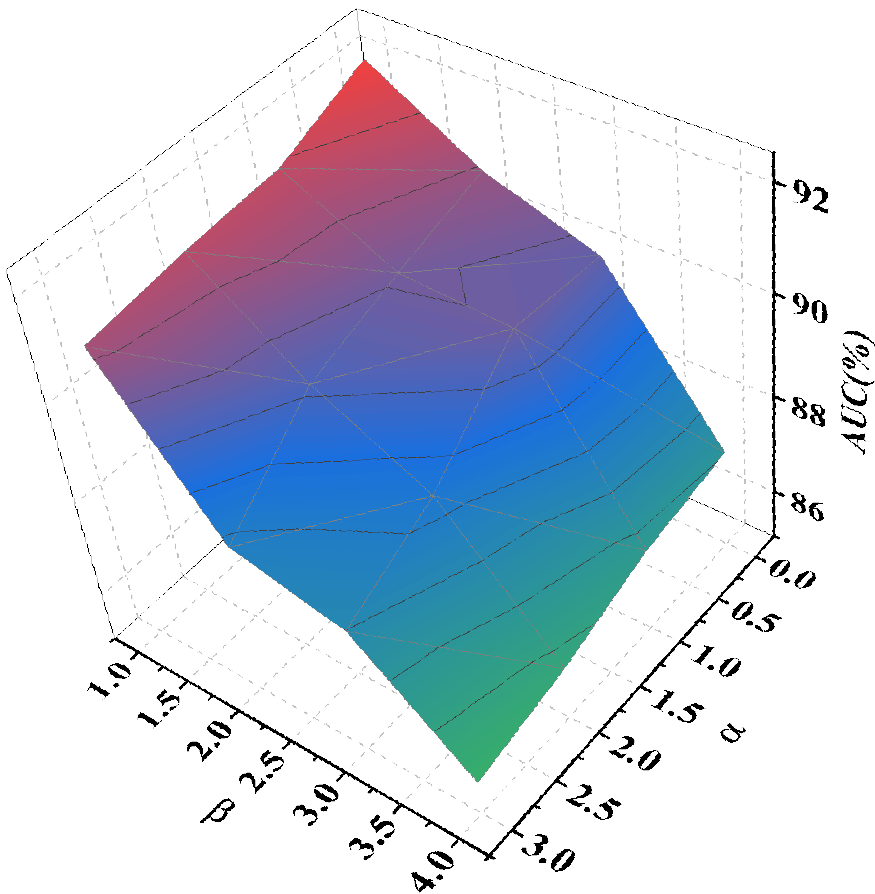} \label{yelppara}}

\subfloat[Amazon]{
		\includegraphics[scale=0.22]{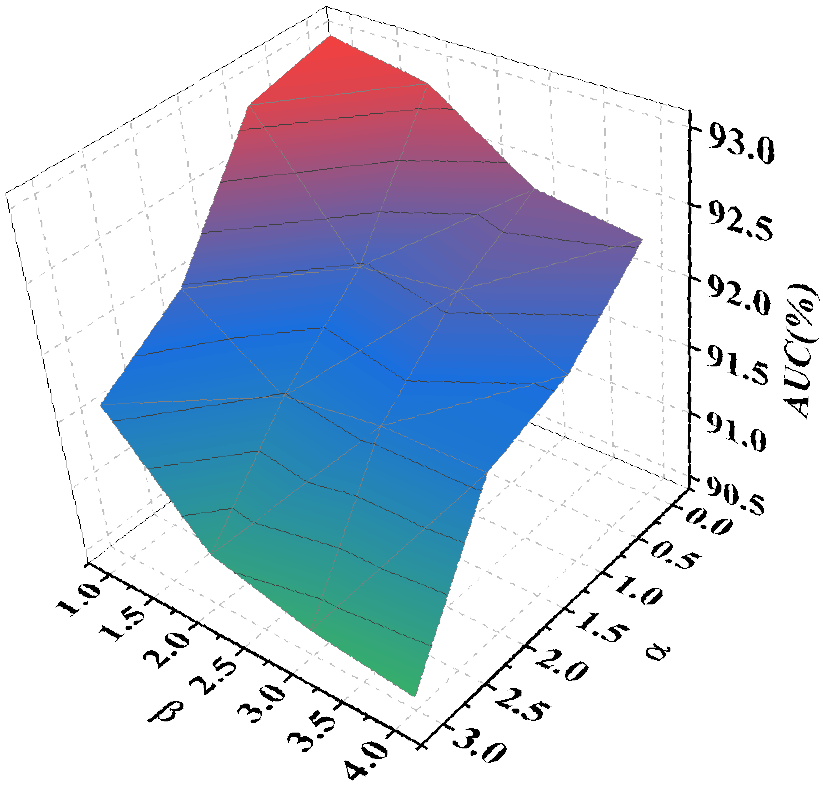} \label{amazonpara}}

\subfloat[FDCompCN]{
		\includegraphics[scale=0.22]{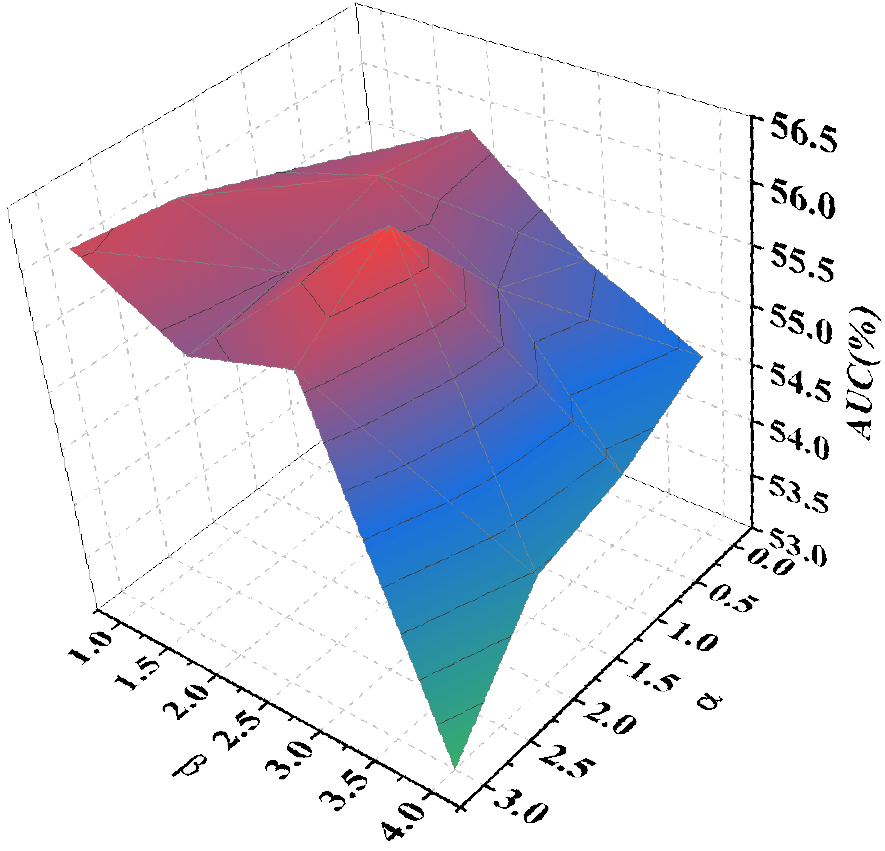} \label{FDpara}}}
\caption{The influence of the selection of parameters of Beta wavelet on the performance on three datasets.}
\label{Model analysis2}
\end{figure*}

\subsection{Model analysis}
In this section, we discuss the selection of key components used in our model. First, we evaluate four different metrics—KL divergence, JS divergence, cosine similarity, and mean square error (MSE)—to compute mutual information.  The comparative results are shown in Figure \ref{Model analysis1}. According to the results, cosine similarity performs the best among the four metrics. Therefore, we choose cosine similarity as the metric for calculating mutual information in  Eq. \ref{I(H;X)all}.

Next, we examine the impact of parameters in Beta wavelet graph filters on the performance of SGNN-IB. Specifically, we vary the parameter $\alpha$ from 0 to 3 and $\beta$ from 1 to 4.  The results, shown in Figure \ref{Model analysis2}, reveal different trends across datasets. 
For the YelpChi and Amazon datasets, performance tends to decrease as the parameter values increase. In contrast, for the FDCompCN dataset, performance improves as the parameters grow larger. This difference is likely due to the sparser structure of the FDCompCN graph. We conclude that high-order wavelet transformations are more effective at capturing detailed information in sparse graphs, while they may introduce excessive noise in denser graphs.

\section{Conclusion}\label{sec:Conclusion}
In this paper, we propose a novel spectral graph network based on information bottleneck (SGNN-IB) for fraud detection in the service networks. SGNN-IB innovatively utilizes an edge classifier to dissect the original service network into heterophilic and homophilic sub-networks. It then applies band-pass graph filters to effectively extract high- and low-frequency service patterns from each subgraph. The framework integrates these signals from multiple relational dimensions to enhance the representation of fraudulent behavior. To improve the robustness and filtering capabilities of the spectral graph network, we introduce an information bottleneck-based learning module. To evaluate the effectiveness and improvements of SGNN-IB, we conduct comprehensive experiments on three publicly available datasets. The results show that our model outperforms existing state-of-the-art methods in terms of detection accuracy.

Despite these promising results, the scalability of our model remains an area for further investigation. Future research will focus on developing more efficient and scalable methods for large-scale fraud detection. Additionally, exploring the potential role of multi-modal information in fraud detection presents an exciting avenue for future work.

\section{Acknowledgment}
This work was supported by the National Natural Science Foundation of China under Grant 72210107001, the Beijing Natural Science Foundation under Grant IS23128, the Fundamental Research Funds for the Central Universities, and by the CAS PIFI International Outstanding Team Project (2024PG0013).

%
%
%
\bibliographystyle{IEEEtran}
\bibliography{ref}

\end{document}